%% file: main.tex
\documentclass[10pt,twocolumn,letterpaper]{article}

\usepackage{cvpr}              

\usepackage{algorithm}
\usepackage{algorithmic}
\usepackage{amsmath}

\usepackage[accsupp]{axessibility}

\input{preamble}
\definecolor{cvprblue}{rgb}{0.21,0.49,0.74}
\usepackage[pagebackref,breaklinks,colorlinks,allcolors=cvprblue]{hyperref}

\def\paperID{*****} 
\def\confName{CVPR}
\def\confYear{2026}

\title{Revisiting Unknowns: Towards Effective and Efficient Open-Set Active Learning}

\author{Chen-Chen Zong, Yu-Qi Chi, Xie-Yang Wang, Yan Cui, Sheng-Jun Huang\thanks{Corresponding Author. }\\
Nanjing University of Aeronautics and Astronautics\\
Nanjing, 211106, China\\
{\tt\small \{chencz,chiyuqi,xieyang,cuiyan,huangsj\}@nuaa.edu.cn}
}

\begin{document}
\maketitle
\input{sec/0_abstract}    
\input{sec/1_intro}
\input{sec/2_related}

\input{sec/3_method}
\input{sec/4_exper}

\input{sec/5_conclusion}

\section*{Acknowledgements}
This work was supported in part by the NSFC (U2441285) and in part by the Interdisciplinary Innovation Fund for Doctoral Students of Nanjing University of Aeronautics and Astronautics.

{
    \small
    \bibliographystyle{ieeenat_fullname}
    \bibliography{main}
}
\appendix
\input{sec/X_suppl}


\end{document}

%% file: sec/0_abstract.tex
\begin{abstract}
Open-set active learning (OSAL) aims to identify informative samples for annotation when unlabeled data may contain previously unseen classes—a common challenge in safety-critical and open-world scenarios. 
Existing approaches typically rely on separately trained open-set detectors, introducing substantial training overhead and overlooking the supervisory value of labeled \texttt{unknowns}\footnote{Here, \texttt{unknowns} denotes samples from unknown (open-set) classes.} for improving known-class learning.
In this paper, we propose E$^2$OAL (Effective and Efficient Open-set Active Learning), a unified and detector-free framework that fully exploits labeled \texttt{unknowns} for both stronger supervision and more reliable querying. E$^2$OAL first uncovers the latent class structure of \texttt{unknowns} through label-guided clustering in a frozen contrastively pre-trained feature space, optimized by a structure-aware F1-product objective. To leverage labeled \texttt{unknowns}, it employs a Dirichlet-calibrated auxiliary head that jointly models known and unknown categories, improving both confidence calibration and known-class discrimination. Building on this, a logit-margin purity score estimates the likelihood of known classes to construct a high-purity candidate pool, while an OSAL-specific informativeness metric prioritizes partially ambiguous yet reliable samples. These components together form a flexible two-stage query strategy with adaptive precision control and minimal hyperparameter sensitivity.
Extensive experiments across multiple OSAL benchmarks demonstrate that E$^2$OAL consistently surpasses state-of-the-art methods in accuracy, efficiency, and query precision, highlighting its effectiveness and practicality for real-world applications. The code is available at \href{https://github.com/chenchenzong/E2OAL}{github.com/chenchenzong/E2OAL}.

\end{abstract}

%% file: sec/1_intro.tex
\section{Introduction}
\label{sec:intro}

Deep learning has achieved remarkable success across a wide range of domains, largely fueled by the availability of large-scale datasets with high-quality annotations~\cite{lecun2015deep,he2016deep,radford2021learning}. However, collecting such annotations is often prohibitively expensive, requiring extensive human effort and domain expertise~\cite{settles2009active,ren2021survey}. These limitations hinder the deployment of deep models in real-world scenarios where labeled data is scarce or costly.

Active learning (AL) mitigates this issue by iteratively selecting a small yet informative subset for annotation~\cite{settles2009active}. Traditional AL methods typically select samples based on model uncertainty~\cite{yang2016active,sharma2017evidence,yoo2019learning}, sample diversity~\cite{sener2017active,agarwal2020contextual,xie2023active}, or combinations thereof~\cite{huang2010active,peng2021active,sun2023evidential}.
Despite their success, most AL methods assume a \textit{closed set}—that all unlabeled samples belong to known classes~\cite{ning2022active}. This assumption rarely holds in practice, especially in safety-critical domains such as autonomous driving~\cite{vieira2022open,perez2024artificial} or medical diagnosis~\cite{budd2021survey,tang2024osal}, where unlabeled data often includes samples from previously unseen classes. 
Under such open-set conditions, conventional AL strategies tend to over-select unknown class samples—mistaking their high uncertainty or feature-level novelty for informativeness—thereby degrading the overall learning process.
These challenges motivate the study of open-set active learning (OSAL), which aims to query informative yet class-consistent samples while properly handling \texttt{unknowns}.

Recent advances in OSAL incorporate surrogate out-of-distribution (OOD) detection mechanisms to guide querying~\cite{du2021contrastive,ning2022active,park2022meta,yang2023not,safaei2024entropic,zong2024bidirectional,zong2025rethinking}. 
These methods typically combine uncertainty or diversity with auxiliary OOD scores to balance informativeness and purity. 
However, they often rely on separately trained OOD detectors, which incur substantial training overhead, and further overlook the potential of labeled \texttt{unknowns} as valuable supervision for enhancing known-class learning. 
This observation calls for a unified, detector-free OSAL framework that can both detect and utilize \texttt{unknowns} to enhance learning and querying.

\begin{figure}[t]
  \centering
   \includegraphics[width=0.96\linewidth]{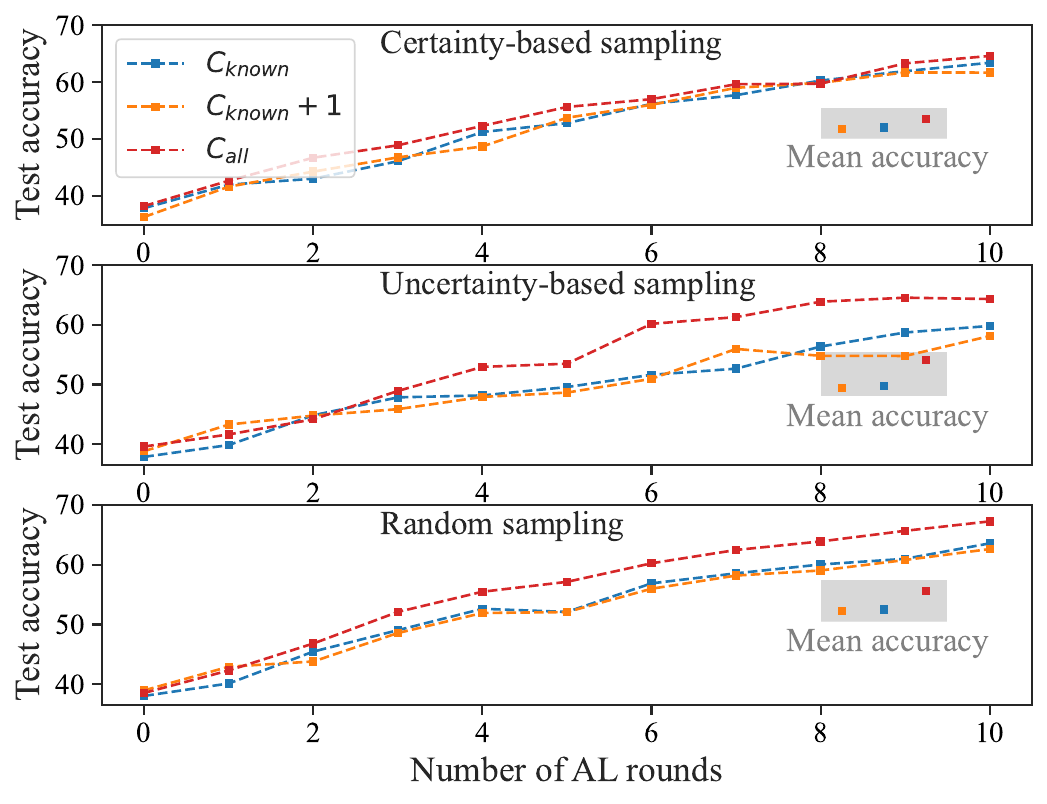}
   \caption{Per-round and mean test accuracy on CIFAR-100 (40 known / 60 unknown) using ResNet-18. $C_{known}$ excludes labeled \texttt{unknowns}, $C_{known{+}1}$ collapses them into a single ``unknown" class, and $C_{all}$ leverages their true labels. $C_{all}$ consistently performs best, suggesting that preserving the latent structure of unknown classes benefits known-class learning.}
   \label{fig:motivation}
\end{figure}

To explore this, we first conduct a pilot study examining whether labeled \texttt{unknowns} benefit known-class learning. 
Assuming access to ground-truth labels for labeled \texttt{unknowns}, we train a dual-head classifier: a primary head optimized for known classes and an auxiliary head jointly trained on both known and unknown categories. 
As shown in Figures~\ref{fig:motivation} and~\ref{fig:motivation_combined} (Appendix), incorporating labeled \texttt{unknowns} into training with their ground-truth labels consistently improves test accuracy. In contrast, merging them into a single ``unknown'' class may hinder performance. 
These results reveal that the latent structure within \texttt{unknowns} provides useful inductive signals. Meanwhile, a well-designed auxiliary head can also implicitly support \texttt{unknowns} detection—without explicit OOD models.

In this paper, we present E$^2$OAL (\textbf{E}ffective and \textbf{E}fficient \textbf{O}pen-set \textbf{A}ctive \textbf{L}earning), a unified, detector-free OSAL framework that converts unknown-class feedback into both stronger supervision and more informative queries. 
At each AL round, E$^2$OAL estimates the number and composition of unknown classes by clustering all labeled samples, including both \texttt{knowns} and \texttt{unknowns}, in a frozen contrastively representation space.
The optimal cluster count is determined by aligning clusters with known labels and maximizing a structure-aware F1-product objective. 
To leverage supervision from labeled \texttt{unknowns}, E$^2$OAL employs a Dirichlet-calibrated auxiliary head that improves known-class discrimination and provides calibrated confidence under open-set conditions. 
For query selection, we introduce a lightweight logit-margin purity score to estimate the likelihood of known classes and a tailored informativeness metric that favors samples with moderate uncertainty while suppressing those with overly ambiguous or confident predictions. 
These scores drive a two-stage query strategy: first constructing a high-purity candidate pool, then selecting the most informative samples for annotation. Crucially, the candidate pool size is dynamically adjusted to meet a target query precision, without introducing additional hyperparameters. 
Our main contributions are summarized as:
\begin{itemize}
    \item We propose E$^2$OAL, a unified and detector-free OSAL framework that transforms unknown-class feedback into both effective supervision and informative querying.

    \item We introduce a label-guided clustering strategy for automatic class estimation within a contrastive feature space, enabling structure-aware discovery of unknown classes.

    \item We develop a Dirichlet-calibrated auxiliary head with an associated logit-margin purity score to enhance known-class learning and provide reliable confidence calibration.

    \item We design an OSAL-specific informativeness metric that favors samples with moderate uncertainty while suppressing overly ambiguous or overconfident ones.

    \item We present a flexible two-stage selection scheme that dynamically adjusts to maintain a fixed target query precision—without introducing additional hyperparameters.

    \item Extensive experiments on multiple OSAL benchmarks show that E$^2$OAL consistently outperforms state-of-the-art methods in accuracy, efficiency, and query precision.
\end{itemize}

%% file: sec/2_related.tex
\section{Related Work}
\label{sec:related}

Active learning (AL) aims to reduce annotation cost by selectively querying the most informative samples while maintaining model performance~\cite{settles2009active,ren2021survey}. 
Classical AL approaches can be broadly categorized into uncertainty-based, diversity-based, and hybrid strategies. 
Uncertainty-based methods query samples with high model uncertainty, measured by entropy~\cite{holub2008entropy}, margin~\cite{balcan2007margin}, confidence scores~\cite{li2006confidence}, or Monte Carlo dropout variance~\cite{gal2017deep}. 
Diversity-based methods instead seek to represent the underlying data distribution through clustering~\cite{nguyen2004active} or coreset selection~\cite{sener2017active}. 
Hybrid methods, such as BADGE~\cite{ash2019deep}, combine both uncertainty and diversity by clustering in the gradient space to balance representativeness and informativeness.

Most conventional AL methods operate under a closed-set assumption, where unlabeled and labeled data share the same label space. 
This assumption, however, fails in open-set active learning (OSAL), where the unlabeled pool may contain out-of-distribution (OOD) or open-set samples irrelevant to the known classes. 
In such cases, standard AL strategies often over-select OOD samples that appear uncertain or diverse yet provide little learning value—much like requesting annotations for unrelated concepts—ultimately degrading model performance.

\begin{figure*}
  \centering
   \includegraphics[width=0.95\linewidth]{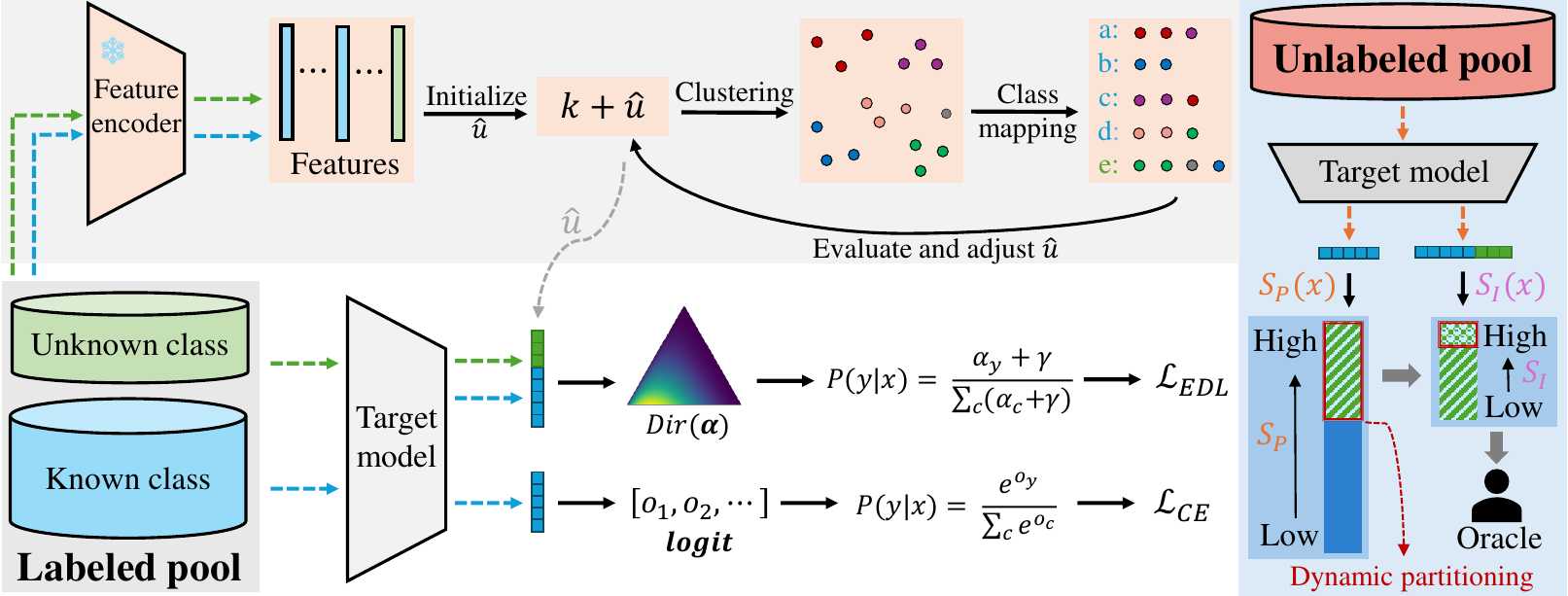}
   \caption{Overview of the proposed E$^2$OAL framework. Each AL round consists of two stages:
(1) Adaptive class estimation and calibration-aware training, where latent unknown classes are discovered via label-guided clustering and incorporated into model learning through Dirichlet-based auxiliary supervision;
(2) Flexible two-stage query selection, where a high-purity candidate pool is first constructed using a purity score guided by a target query precision, followed by informativeness-driven sample selection.}
   \label{fig:framework}
\end{figure*}

To address this, various OSAL methods have been developed. 
LfOSA~\cite{ning2022active} trains an auxiliary detector to distinguish known classes, guiding queries using maximum activation values. 
MQNet~\cite{park2022meta} adopts a meta-learning framework that balances purity and informativeness via a multilayer perceptron. 
More recent works such as EOAL~\cite{safaei2024entropic}, BUAL~\cite{zong2024bidirectional}, and EAOA~\cite{zong2025rethinking} jointly assess purity and informativeness within a detector-based paradigm: 
EOAL defines separate entropy scores for known and unknown classes; 
BUAL leverages bidirectional uncertainty to refine selection boundaries; 
and EAOA integrates epistemic and aleatoric uncertainty for improved sample evaluation.

Despite recent progress, most OSAL methods still depend on separately trained detector networks, which add notable computational burden.
In addition, they fail to fully exploit labeled \texttt{unknowns}—annotations that could provide valuable supervision for improving known-class discrimination.
These limitations motivate our work, which revisits the role of labeled \texttt{unknowns} and introduces a unified, detector-free OSAL framework that converts them into effective supervision and reliable query guidance.

%% file: sec/3_method.tex
\section{The Proposed Approach}
\label{sec:method}

In this section, we present E$^2$OAL, a unified and detector-free framework for open-set active learning (OSAL) in multi-class image classification, where the unlabeled pool may contain samples from previously unseen classes.
An overview of the framework is illustrated in Figure~\ref{fig:framework}, with detailed pseudocode provided in Algorithm~\ref{alg:active_learning_query} (Appendix).

\subsection{Preliminaries}

\textbf{Notations.} 
We consider an OSAL setting where the unlabeled pool $\mathcal{D}_U$ contains both known and unknown classes.
Let $\mathcal{C}_k$ and $\mathcal{C}_u$ denote the sets of known and unknown classes, with $|\mathcal{C}_k|=k$, $|\mathcal{C}_u|=u$, and $\mathcal{C}_k \cap \mathcal{C}_u = \emptyset$.
The labeled set $\mathcal{D}_L$ initially includes only samples from $\mathcal{C}_k$, while $\mathcal{D}_U$ covers $\mathcal{C}_k \cup \mathcal{C}_u$, with $|\mathcal{D}_L| \ll |\mathcal{D}_U|$.
At each active learning (AL) round, a batch $\mathcal{B} \subset \mathcal{D}_U$ is queried and annotated.
If a sample belongs to $\mathcal{C}_k$, its true class label is provided; otherwise, it receives a single ``unknown'' label, without revealing its actual class in $\mathcal{C}_u$.
The newly labeled batch is added to $\mathcal{D}_L$, resulting in $\mathcal{D}_L = \mathcal{D}_L^{kno} \cup \mathcal{D}_L^{unk}$ for known and unknown subsets, respectively.
The observed query precision is $\bar{p}^* = |\mathcal{B}^{kno}| / |\mathcal{B}|$.
Our goal is to train a classifier $f_\theta$ that generalizes well on $\mathcal{C}_k$ with minimal labeling cost, even under contamination from \texttt{unknowns}.

\textbf{Overview.} 
Each AL round in E$^2$OAL proceeds in two sequential stages: first, it uncovers and exploits the latent class structure of labeled \texttt{unknowns} to strengthen model training; then, it selects informative yet likely-known instances through an adaptive query strategy.

\textit{In stage 1}, E$^2$OAL performs label-guided clustering on $\mathcal{D}_L$ within a frozen contrastive representation space (e.g., from CLIP~\cite{radford2021learning}, MoCo~\cite{he2020momentum}, or SimCLR~\cite{chen2020simple}) to reveal latent unknown-class structures.
A simple structure-aware F1-product objective determines the optimal cluster configuration, enabling adaptive class discovery.
The resulting pseudo-clusters are treated as auxiliary classes for an additional classification head, trained jointly with the primary head using a Dirichlet-based calibration loss.
This joint training improves known-class discrimination and yields calibrated confidence estimates.
\textit{In stage 2}, E$^2$OAL estimates a logit-margin purity score for each unlabeled sample to measure its likelihood of belonging to known classes.
Samples surpassing an adaptively adjusted purity margin, set to maintain a target query precision, form a high-purity candidate pool.
Within this pool, a tailored informativeness metric evaluates each sample’s utility under open-set uncertainty, prioritizing those that are both reliable and informative.
By combining these mechanisms, E$^2$OAL adaptively selects high-quality queries and achieves efficient, detector-free OSAL without additional hyperparameter tuning.

\subsection{Adaptive Class Estimation}

To utilize labeled \texttt{unknowns} effectively, E$^2$OAL estimates the latent class structure of $\mathcal{D}_L^{unk}$, including both the number and composition of unknown categories.
Unlike traditional clustering methods that assume a fixed number of classes, E$^2$OAL operates under the open-set assumption, where neither the number nor the granularity of unknown classes is known a priori.

To capture latent relationships among samples, we operate in a semantically rich and noise-robust contrastively pretrained feature space. Here, we leverage frozen representations from CLIP~\cite{radford2021learning} \footnote{Our experiments show that MoCo and CLIP achieve similar performance, while CLIP can be directly used without additional training.}, a pre-trained vision-language model known for providing high-quality, task-agnostic embeddings without additional fine-tuning.
Within this feature space, we perform clustering on the labeled set $\mathcal{D}_L$.
For each candidate number of clusters $\hat{u} \in \{k+1, k+2, \dots, \hat{u}_{\max}\}$, clustering quality is evaluated by one-to-one aligning predicted assignments with ground-truth labels (covering $k$ known classes and a unified ``unknown'' class) using Hungarian algorithm~\cite{kuhn1955hungarian}.
Assignments unmatched to known classes are treated as unknown categories.
The clustering score is defined as the product of per-class F1 scores, jointly capturing alignment and structural fidelity.

The choice of $\hat{u}$ has a strong impact on clustering quality~\cite{vaze2022generalized}: underestimating $\hat{u}$ may merge distinct known classes, while overestimating it can fragment coherent ones; both reduce the F1 score and hence the product.
To determine the optimal $\hat{u}$ efficiently, we employ a ternary search to maximize the F1-product score, thereby identifying the best cluster configuration. Each labeled unknown sample is then assigned to its nearest inferred cluster, which serves as a proxy category for downstream training.

The detailed pseudocode is provided in Algorithm~\ref{alg:ternary_search_estimation}.

\begin{algorithm}
\caption{The adaptive class estimation algorithm}
\label{alg:ternary_search_estimation}
\textbf{Input:} Labeled data pool $\mathcal{D}_L = \{(x_i, y_i)\}$, known class count $k$, and upper limit $\hat{u}_{max}$ \\
\textbf{Output:} Estimated number of unknown classes $\hat{u}$

\begin{algorithmic}[1]
    \STATE Extract CLIP features $\{f_i\}$ for all $x_i \in \mathcal{D}_L$

    \STATE \textit{\# Ternary search for optimal unknown class count}
    \STATE Initialize bounds: $l \gets k + 1$, $r \gets \hat{u}_{max}$
    \WHILE{$r - l > 2$}
        \STATE $m_1 \gets \lfloor l + \frac{r - l}{3} \rfloor$, \quad $m_2 \gets \lfloor r - \frac{r - l}{3} \rfloor$
        \STATE $s_{m_1} \gets$ \textsc{Evaluate}($m_1$, $\{f_i\}$, $k$)
        \STATE $s_{m_2} \gets$ \textsc{Evaluate}($m_2$, $\{f_i\}$, $k$)
        \IF{$s_{m_1} < s_{m_2}$}
            \STATE $l \gets m_1$
        \ELSE
            \STATE $r \gets m_2$
        \ENDIF
    \ENDWHILE
    \STATE $\hat{u} \gets \arg\max_{m \in \{l, l+1, \dots, r\}}$ \textsc{Evaluate}($m$, $\{f_i\}$, $k$)
\end{algorithmic}
\textbf{Function} \textsc{Evaluate}($m$, $\{f_i\}$, $k$):

\begin{algorithmic}[1]
    \STATE $C \gets k + m$
    \STATE Perform K-Means clustering on $\{f_i\}$ into $C$ clusters
    \STATE Match the $k+1$ clusters to $k+1$ classes—including all $k$ known classes and a unified unknown class—using the Hungarian algorithm~\cite{kuhn1955hungarian}
    \STATE Assign the remaining $(C - (k+1))$ clusters to the unknown class as well
    \STATE Compute class-wise F1 scores $\{\text{F1}_c\}_{c=1}^{k+1}$
    \STATE \textbf{return}~$\prod_{c=1}^{k+1} \text{F1}_c$
\end{algorithmic}
\end{algorithm}

\subsection{Dirichlet-Based Calibration}
While incorporating labeled \texttt{unknowns} improves generalization, reliable confidence estimation in open-set settings remains challenging. Standard softmax classifiers are notoriously overconfident~\cite{sensoy2018evidential,grabinski2022robust,zong2024dirichlet}, especially for outliers, because the softmax function is translation-invariant—adding or subtracting a constant from all logits does not alter the predicted probabilities. For example, logits $[0, 5, 0, 0]$ and $[-5, 0, -5, -5]$ both yield a softmax confidence of roughly $0.88$ for the second class, even though the latter encodes far weaker semantic evidence. This results in misleadingly high confidence for semantically ambiguous or abnormal inputs, which is particularly detrimental in OSAL, where confidence estimation plays a critical role.

To alleviate this overconfidence, we adopt a translation-aware variant of softmax that explicitly breaks translation invariance by introducing an additive constant. Given logits $\boldsymbol{o} = [o_1, o_2, \dots, o_{k+\hat{u}}]$ for a sample $x$, the calibrated probability for class $y$ is:
\begin{equation}\label{eq1}
P(y|x) = \frac{e^{o_{y}} + \gamma}{\sum_{c=1}^{k+\hat{u}} (e^{o_{c}} + \gamma)},
\end{equation}
where $\gamma > 0$ is a calibration constant. This modification ensures that predicted probabilities reflect both relative logit magnitudes and absolute evidence. The additive term $\gamma$ serves as a smoothing factor, suppressing overconfidence for low-evidence samples. For instance, with $\gamma = 1$, the probabilities for the above logits decrease from $0.88$ to approximately $0.60$ and $0.38$, respectively, improving confidence separation between confident and ambiguous cases.


However, directly optimizing this formulation with the commonly adopted cross-entropy (CE) loss introduces a gradient mismatch: the additive constant enlarges the logit margin required for confident predictions, thereby requiring gradient compensation to maintain effective optimization.

A straightforward remedy is to impose auxiliary regularization (e.g., $\ell_2$ or KL divergence) to promote calibrated outputs. However, such rigid constraints often conflict with CE loss, leading to unstable optimization. To address this, we adopt Evidential Deep Learning (EDL)~\cite{sensoy2018evidential,zong2024dirichlet,li2024hyper}, which models predictive probabilities as Dirichlet distributions and provides a principled, soft calibration mechanism naturally aligned with the modified softmax.

In EDL, the predictive distribution $\boldsymbol{p} = [p_1, \dots, p_{k+\hat{u}}]$ is drawn from a Dirichlet distribution $\mathrm{Dir}(\boldsymbol{\alpha})$, parameterized by $\boldsymbol{\alpha} = [\alpha_1, \dots, \alpha_{k+\hat{u}}]$. The expected class probability is:
\begin{equation}\label{eq2}
P(y|x) = \mathbb{E}_{\boldsymbol{p} \sim \mathrm{Dir}(\boldsymbol{\alpha})}[p_y] = \frac{\alpha_y}{\sum_{c=1}^{k + \hat{u}} \alpha_c}.
\end{equation}
We define $\boldsymbol{\alpha}$ via the model’s logits as $\boldsymbol{\alpha} = \frac{g(\boldsymbol{o})}{\gamma} + 1$, where $g(\cdot)$ is a non-negative activation (e.g., exponential) and $\gamma$ is the calibration constant in Eq.~\eqref{eq1}, which can be implicitly learned by the network during training. The term $\boldsymbol{e} = g(\boldsymbol{o})$ represents the \textit{evidence}—the model’s support for each class. Substituting $\boldsymbol{\alpha} = \frac{g(\boldsymbol{o})}{\gamma} + 1$ into Eq.~\eqref{eq2} with $g(\cdot) = \exp(\cdot)$ recovers the translation-aware softmax in Eq.~\eqref{eq1}.

Training is guided by a composite objective combining predictive accuracy and distributional regularization. The first term, a negative log-likelihood (NLL) loss, encourages high expected confidence on the true label:
\begin{equation}\label{eq3}
\mathcal{L}_{\mathrm{NLL}} = - \log P(y|x) = - \log \frac{\alpha_{y}}{\sum_{c=1}^{k+\hat{u}} \alpha_{c}}.
\end{equation}
The second term, a KL divergence, penalizes excessive evidence for incorrect classes by regularizing the output Dirichlet toward a non-informative uniform prior:
\begin{equation}\label{eq4}
\mathcal{L}_{\mathrm{KL}} = \mathrm{KL}\big(\mathrm{Dir}(\tilde{\boldsymbol{\alpha}}) \parallel \mathrm{Dir}(\mathbf{1})\big),
\end{equation}
where $\tilde{\boldsymbol{\alpha}} = \boldsymbol{y} + (1 - \boldsymbol{y}) \odot \boldsymbol{\alpha}$, $\boldsymbol{y}$ is the one-hot label encoding, $\mathbf{1}$ is a vector of ones, and $\odot$ denotes element-wise multiplication. The full expansion of Eq.~\eqref{eq4} is provided in the Appendix (\cref{fe_eq4}).

Finally, the overall training loss is defined as:
\begin{equation}\label{eq5}
\mathcal{L} = \mathcal{L}_{\mathrm{CE}} + \mathcal{L}_{\mathrm{EDL}} = \mathcal{L}_{\mathrm{CE}} + (\mathcal{L}_{\mathrm{NLL}} + \mathcal{L}_{\mathrm{KL}}),
\end{equation}
where the auxiliary head is optimized over $k + \hat{u}$ categories using $\mathcal{L}_{\mathrm{NLL}}$ and $\mathcal{L}_{\mathrm{KL}}$, while the primary head only focuses on the known classes via $\mathcal{L}_{\mathrm{CE}}$.

\subsection{Flexible Information-Purity Sampling}
E$^2$OAL performs query selection by jointly balancing two key factors: class purity and sample informativeness. High informativeness promotes label efficiency, while high purity ensures that most queries originate from known classes, thereby minimizing contamination from \texttt{unknowns}.

A fundamental challenge in OSAL is to identify unlabeled samples that are most likely to belong to known classes. We address this by introducing a lightweight logit-margin purity score computed from the auxiliary head’s calibrated logits.  
Under Dirichlet-based calibration, each logit encodes class-specific evidence: positive values indicate support, with larger magnitudes reflecting higher confidence, while negative values indicate rejection.  
For an unlabeled sample $x \in \mathcal{D}_U$, let $o_{(1)}$ and $o_{(2)}$ denote the highest logits among known and unknown classes, respectively. The purity score is defined as:
\begin{equation}\label{eq:purity}
S_{\text{purity}}(x) = o_{(1)} - o_{(2)} = \max_{c \in \mathcal{C}_k} o_c - \max_{c \in \mathcal{C}_{\hat{u}}} o_c.
\end{equation}
A higher $S_{\text{purity}}(x)$ indicates stronger evidence for known-class membership.  
Unlike softmax-based confidence margins, this metric explicitly measures evidence separation, providing enhanced robustness under open-set conditions.

\begin{figure*}[!ht]
\centering
   \includegraphics[width=1.0\linewidth]{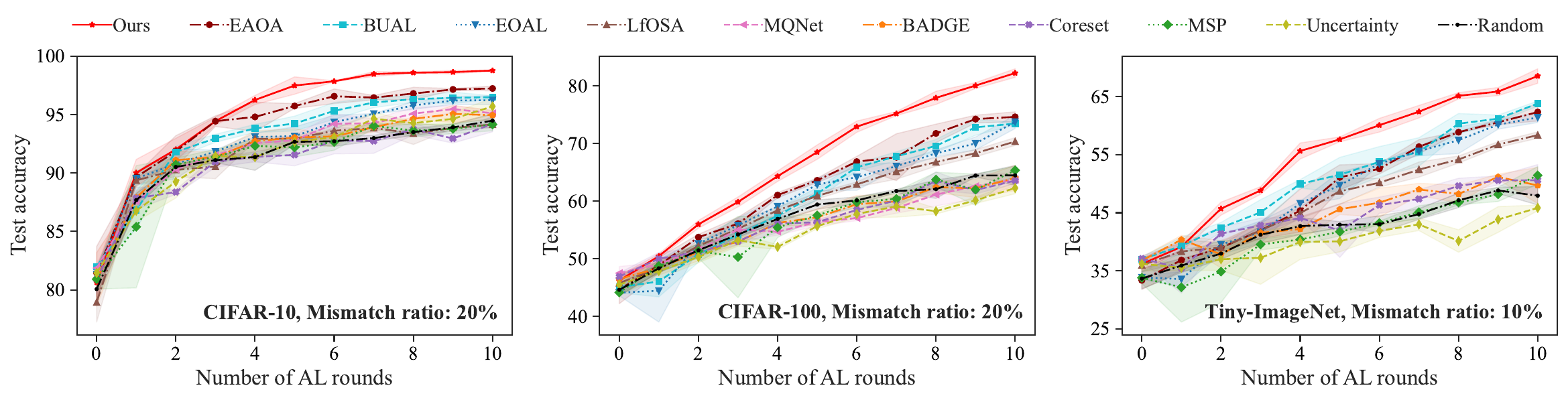}
   \includegraphics[width=1.0\linewidth]{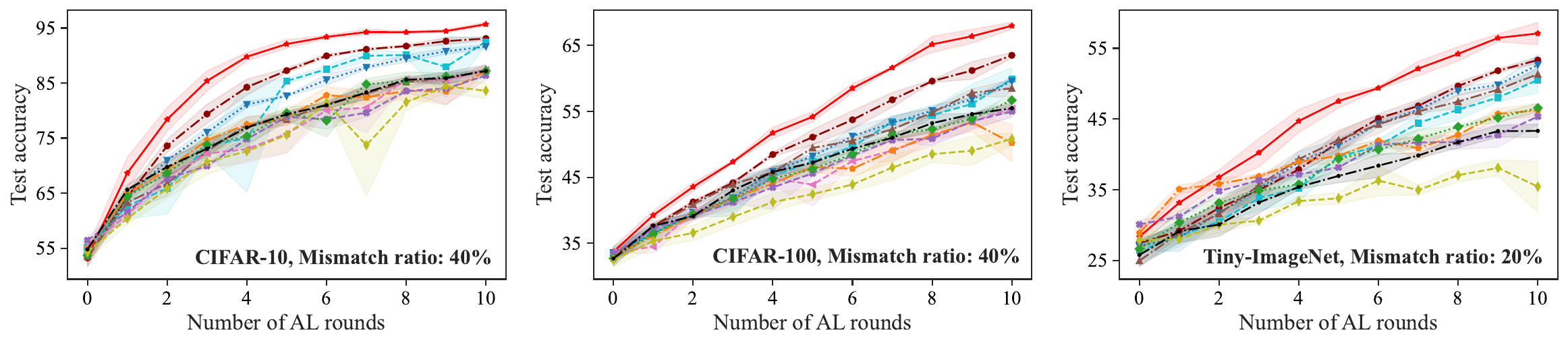}
\caption{Test accuracy across AL rounds under varying mismatch ratios on CIFAR-10/100 and Tiny-ImageNet. }
	\label{fig:acc_curve}
\end{figure*}

E$^2$OAL further prioritizes samples that are both likely from known classes and highly informative.  
To quantify informativeness under open-set conditions, we propose a Jensen–Shannon (JS) divergence–based metric between the primary head’s output and two reference distributions:
\begin{equation}\label{eq:informativeness}
S_{\text{info}}(x) = \mathrm{JS}\big(\mathbf{p} \parallel \mathbf{u}\big) \cdot \mathrm{JS}\big(\mathbf{p} \parallel \mathbf{p}^{\mathrm{max}}\big),
\end{equation}
where $\mathbf{p}$ is the predicted probability vector from the primary head, $\mathbf{u}$ is the uniform distribution, and $\mathbf{p}^{\mathrm{max}}$ is the one-hot encoding of the most confident class in $\mathbf{p}$.  
This formulation emphasizes moderately uncertain samples, which diverge from both uniform and peaked distributions, thus favoring ambiguous yet informative instances while avoiding outliers or trivial predictions.

E$^2$OAL integrates the above two metrics into a flexible two-stage selection scheme calibrated to meet a fixed target query precision $p^*$.
\textit{In stage 1}, a high-purity candidate pool $\mathcal{C}_{\text{pool}}$ is constructed by fitting a three-component Gaussian Mixture Model (GMM)~\cite{permuter2006study} to all $S_{\text{purity}}(x)$ values in $\mathcal{D}_U$.  
Each GMM component corresponds to a distinct purity regime: high (known-class-like), low (unknown-class-like), and intermediate (ambiguous).  
Samples are ranked by their posterior probability of belonging to the high-purity component, and $\mathcal{C}_{\text{pool}}$ is expanded until the mean posterior probability of its bottom-$|\mathcal{B}|$ entries matches a calibrated query precision $\hat{p}^*$.
Notably, due to systematic estimation bias, we use a calibrated query precision $\hat{p}^*$ instead of directly applying the target precision $p^*$.
To compensate for such systematic shifts, the calibrated precision $\hat{p}^*$ is adaptively updated using the observed precision $\bar{p}^*$ from the previous round and the target precision $p^*$ as follows:
\begin{equation}
\hat{p}^*_{t+1} {=} 
\begin{cases}
\max\big(\min(\hat{p}^*_t + (p^* - \bar{p}_t^*), 0), 1\big) & \text{if } t {>} 0, \\
p^* & \text{if } t {=} 0.
\end{cases}
\end{equation} 
This adaptive mechanism eliminates manual threshold tuning and ensures stable query precision across rounds.

\textit{In stage 2}, the top-$|\mathcal{B}|$ samples from $\mathcal{C}_{\text{pool}}$ are selected according to $S_{\text{info}}(x)$.  
This two-stage strategy achieves a principled balance between uncertainty-driven exploration and label-purity control, enabling robust and adaptive querying.

%% file: sec/4_exper.tex
\section{Experiments}

\textbf{Datasets.}
We evaluate our method E$^2$OAL on three standard image classification benchmarks: CIFAR-10~\cite{krizhevsky2009learning}, CIFAR-100~\cite{krizhevsky2009learning}, and Tiny-ImageNet~\cite{yao2015tiny}, which contain 10, 100, and 200 classes, respectively.
To simulate open-set active learning (OSAL) conditions, a subset of classes is designated as ``known'', with the remainder treated as ``unknown''.  
The proportion of known classes, referred to as the \emph{mismatch ratio}, is set to 20\%, 30\%, and 40\% for CIFAR-10/100, and 10\%, 15\%, and 20\% for Tiny-ImageNet, introducing varying levels of open-set difficulty.

\textbf{Training Setup.}
For each dataset, the labeled pool is initialized by randomly sampling 1\% of known-class instances for CIFAR-10 and 8\% for CIFAR-100 and Tiny-ImageNet.  
We perform 10 active learning (AL) rounds, querying 1,500 samples per round. 
All models adopt a ResNet-50~\cite{he2016deep} backbone and are trained with SGD (momentum 0.9, weight decay $5\times10^{-4}$, batch size 128) for 200 epochs, starting with a learning rate of 0.01 decayed by 0.1 every 60 epochs.  
Experiments are performed on NVIDIA RTX 3090 GPUs, and all results are averaged over three runs with random seeds \{1, 2, 3\}.  
The target query precision is fixed at $p^*=0.6$, consistent with prior work~\cite{zong2025rethinking}, while the estimated upper bound of unknown classes is set to $\hat{u}_{\max} = 1000$.

\textbf{Compared Methods.}
We compare E$^2$OAL against representative methods from four categories:
(1) Random sampling;
(2) Informativeness-based approaches, including Uncertainty~\cite{li2006confidence}, Coreset~\cite{sener2017active}, and BADGE~\cite{ash2019deep};
(3) Purity-based methods, such as MSP~\cite{hendrycks2016baseline} and LfOSA~\cite{ning2022active};
(4) Hybrid approaches, including MQNet~\cite{park2022meta}, EOAL~\cite{safaei2024entropic}, BUAL~\cite{zong2024bidirectional}, and EAOA~\cite{zong2025rethinking}, which jointly consider informativeness and purity.
All methods follow identical training and evaluation protocols for fair comparison.

\begin{figure*}[!ht]
\centering
   \includegraphics[width=0.955\linewidth]{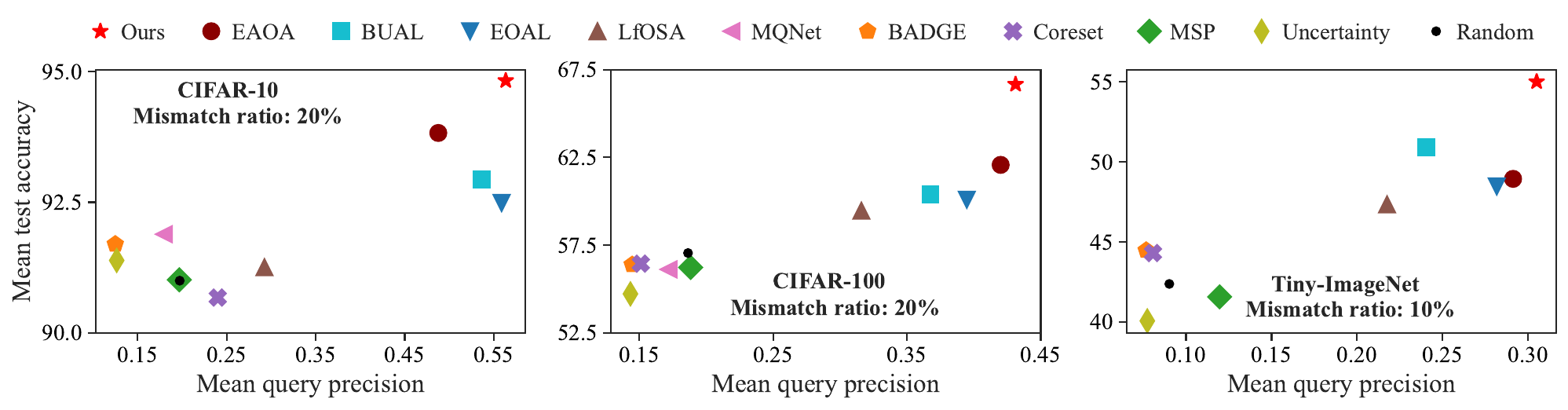}
   \includegraphics[width=0.955\linewidth]{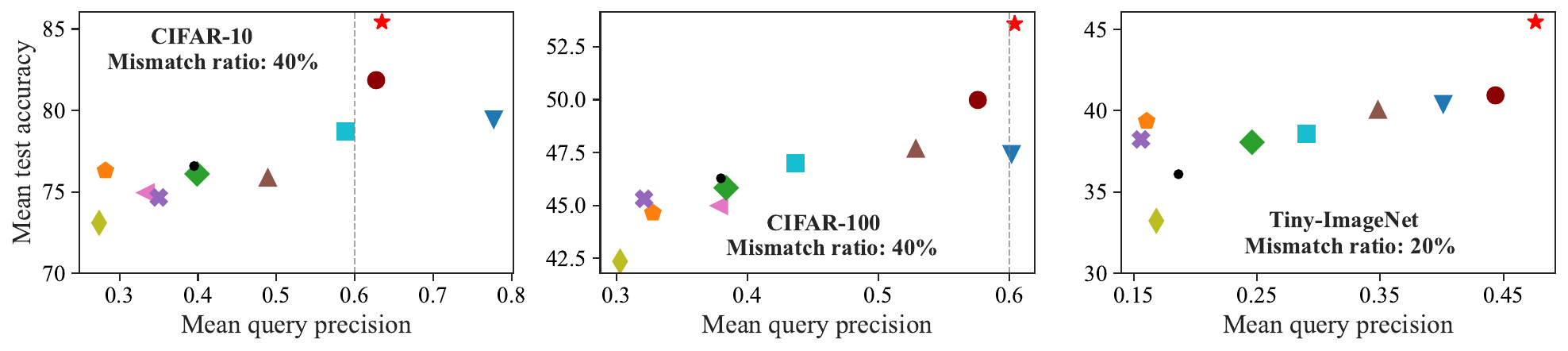}
\caption{Mean query precision vs. mean test accuracy across rounds under varying mismatch ratios on CIFAR-10/100 and Tiny-ImageNet.
}
	\label{fig:query}
\end{figure*}

\subsection{Performance Comparison}
We assess the classification performance and query effectiveness of E$^2$OAL on CIFAR-10, CIFAR-100, and Tiny-ImageNet.
Figure~\ref{fig:acc_curve} illustrates the progression of test accuracy across AL rounds, while Figure~\ref{fig:query} presents the relationship between mean query precision and mean test accuracy.
Due to space constraints, results for intermediate mismatch ratios are provided in Appendix Figures~\ref{fig:acc_curve_appendix} and~\ref{fig:query_appendix}.

As shown in Figures~\ref{fig:acc_curve} and~\ref{fig:acc_curve_appendix} (Appendix), E$^2$OAL consistently surpasses all baselines across varying mismatch ratios and datasets, maintaining clear advantages throughout the entire query process.  
These results demonstrate the robustness and effectiveness of our framework. 
Among the baselines, hybrid approaches such as EAOA, BUAL, and EOAL outperform single-factor methods by jointly balancing purity and informativeness. However, their reliance on separate detectors and limited use of labeled \texttt{unknowns} restricts overall performance.  
In contrast, MQNet combines both factors but suffers from inaccurate purity estimation and also fails to effectively leverage labeled \texttt{unknowns}.

Figures~\ref{fig:query} and~\ref{fig:query_appendix} (Appendix) further show that E$^2$OAL consistently lies in the top-right region, achieving both high query precision and superior test accuracy.  
Compared with hybrid baselines (e.g., EAOA, BUAL, EOAL), E$^2$OAL achieves a better balance between precision and informativeness, attributed to its adaptive sampling and effective use of labeled \texttt{unknowns}.  
Moreover, E$^2$OAL maintains query precision close to the target $p^*=0.6$, demonstrating better control than EAOA, which is also guided by a target precision.  
In contrast, informativeness-only methods (e.g., Uncertainty, Coreset, BADGE) tend to oversample \texttt{unknowns}, while purity-only methods (e.g., MSP, LfOSA) often query redundant known-class samples.  
By harmonizing these two objectives, E$^2$OAL enables more efficient querying and improved generalization.

In addition, Figure~\ref{fig:query_curve_appendix} (Appendix) further illustrates how query precision evolves over AL rounds. 
We observe that even in the early stages, our method consistently maintains high query precision, whereas the previous state-of-the-art method, EAOA, exhibits suboptimal precision and notable fluctuations. 
This instability originates from the inherent limitation of its adaptive strategy, which updates with a fixed step size and struggles to converge quickly to the optimal value. 
In contrast, our method introduces no additional hyperparameters and achieves stable, high-precision querying from the outset, demonstrating clear superiority.

\subsection{Ablation Studies}

\textbf{Impact of Each Module.}
We first examine a variant where the target classifier is trained independently without using labeled \texttt{unknowns}, as in prior methods. As shown in Table~\ref{tab:accuracy_comparison} (see Table~\ref{tab:final-acc} in Appendix for complete results), our method still achieves consistent performance gains, particularly on the challenging Tiny-ImageNet, confirming the effectiveness of the proposed adaptive sampling strategy.

\begin{table}[t]
\centering
\small
\begin{tabular}{lcccc}
\toprule
Method & CIFAR-10 & CIFAR-100 & Tiny-ImageNet \\
\midrule
Ours*      & \textbf{95.94} & \textbf{67.54} & \textbf{60.44} \\
EAOA    & 95.88  & 67.14 & 57.31 \\
BUAL & 95.04 & 63.73 & 56.09 \\
EOAL & 93.64 & 63.69 & 56.13 \\
\bottomrule
\end{tabular}
\caption{Final-round test accuracy (\%) under fixed mismatch ratios (30\% for CIFAR-10/100 and 15\% for Tiny-ImageNet).
``Ours*'' denotes a variant trained without leveraging labeled \texttt{unknowns}.}
\label{tab:accuracy_comparison}
\end{table}

\begin{table}[t]
\centering
\small
\begin{tabular}{lccc}
\toprule
Method & CIFAR-10 & CIFAR-100 & Tiny-ImageNet \\
\midrule
$\mathcal{L}_{\mathrm{EDL}}$    & \textbf{9495} & \textbf{7934} & \textbf{5844} \\
$\mathcal{L}_{\mathrm{CE}}$     & 9394 & 7814 & 5661 \\
\bottomrule
\end{tabular}
\caption{Purity comparison under the most challenging mismatch ratios: 20\% for CIFAR-10/100 and 10\% for Tiny-ImageNet. 
Sampling by Eq.~\eqref{eq:purity} with CE and EDL losses on the auxiliary head, evaluated by the total number of queried known-class samples.
}
\label{tab:purity_comparison}
\end{table}

We then separately analyze the contributions of the purity and informativeness components. 
As presented in Table~\ref{tab:purity_comparison}, our Dirichlet-based calibration significantly enhances known-class discrimination, yielding a larger number of known samples among queried instances and thereby improving overall purity.
For informativeness, Table~\ref{tab:informativeness_comparison} shows that our moderately uncertainty-based metric consistently outperforms its counterpart in EAOA—the previous state of the art—achieving higher test accuracy and demonstrating stronger capability in identifying informative samples.

\begin{table}[t]
\centering
\small
\begin{tabular}{lccc}
\toprule
Method & CIFAR-10 & CIFAR-100 & Tiny-ImageNet \\
\midrule
Ours      & \textbf{95.95} & \textbf{65.73} & \textbf{48.63} \\
EAOA     & 94.90 & 61.95 & 44.73 \\
Uncertainty     & 95.37 & 61.28 & 44.73 \\
\bottomrule
\end{tabular}
\caption{Informativeness comparison under the most challenging mismatch ratios: 20\% for CIFAR-10/100 and 10\% for Tiny-ImageNet. 
Final test accuracy (\%) when retaining only the informativeness component from ``Ours", ``EAOA", and ``Uncertainty".
}
\label{tab:informativeness_comparison}
\end{table}

\begin{table}[t]
\centering
\small
\begin{tabular}{lccc}
\toprule
Method & CIFAR-10 & CIFAR-100 & Tiny-ImageNet \\
\midrule
Ours & \textbf{97.52} & \textbf{72.10} & \textbf{64.02}\\ 
w/o ClassExp & 97.17 & 70.73 & 62.67 \\ 
$S_{\text{purity}}$  & 96.73    & 72.00 & 61.93 \\
$S_{\text{info}}$     & 96.00 & 68.20 & 57.60 \\
\bottomrule
\end{tabular}
\caption{Final-round test accuracy (\%) of different ablation variants under intermediate mismatch ratios on CIFAR-10/100 and Tiny-ImageNet. ``$S_{\text{info}}$" and ``$S_{\text{purity}}$" denote sampling based solely on informativeness or purity. ``w/o ClassExp" disables class expansion by collapsing all labeled \texttt{unknowns} into a single class.}
\label{tab:component_analysis}
\end{table}

\begin{table}[t]
\centering
\small
\begin{tabular}{lcccc}
\toprule
Mismatch ratio & $p^*=0.4$ & $0.5$ & $0.6$ & $0.7$ \\ 
\midrule
$20\%$ & 81.47 & 81.67 & \textbf{82.20} & 82.18 \\
$30\%$ & 71.90 & 72.96 & 72.10 & \textbf{73.17} \\
$40\%$ & 66.68 & \textbf{68.03} & 67.98 & 66.78 \\
\bottomrule
\end{tabular}
\caption{Final-round test accuracy (\%) on CIFAR-100 under varying target query precisions and mismatch ratios.}
\label{tab:precision_sensitivity}
\end{table}

\begin{figure}[!ht]
  \centering
   \includegraphics[width=1.\linewidth]{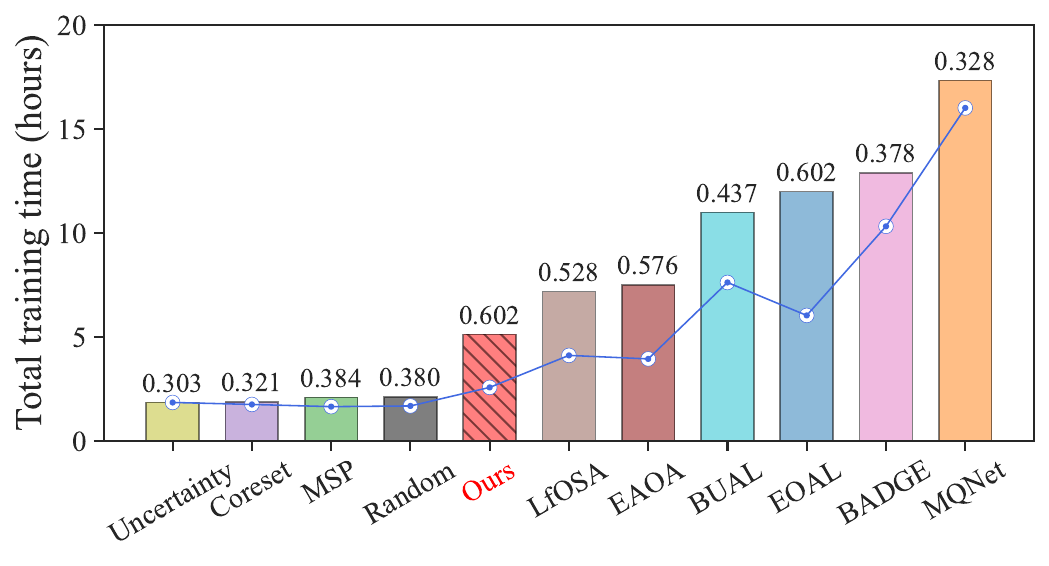}

   \caption{Total training time (hours) on CIFAR-100 under a 40\% mismatch ratio. Bars indicate actual training time with the average query precision annotated; the dashed line shows the approximate projection assuming a linear relationship between query precision and time, aligned to the precision level of ``Uncertainty".}
   \label{fig:training_time}
\end{figure}

Finally, Table~\ref{tab:component_analysis} evaluates the contribution of each component.  
Using only or disabling any single module notably reduces performance across datasets, confirming that each contributes complementarily to the overall framework.

\textbf{Sensitivity to Target Precision $p^*$.}
In the main experiments, the target query precision $p^*$ is fixed at 0.6, following EAOA. 
To examine the sensitivity of E$^2$OAL to this parameter, we further evaluate it under $p^* \in \{0.4,\ 0.5,\ 0.6,\ 0.7\}$. 
As shown in Table~\ref{tab:precision_sensitivity}, although the optimal value may vary slightly with the mismatch ratio, E$^2$OAL consistently achieves strong performance across all settings, demonstrating its robustness and adaptability.

\textbf{Training Efficiency.}
Figure~\ref{fig:training_time} presents a comparison of the overall and equivalent training times across different methods.
E$^2$OAL achieves high training efficiency, with its equivalent time comparable to lightweight baselines such as Random, MSP, Coreset, and Uncertainty. By removing the need for separate detectors and effectively utilizing labeled \texttt{unknowns}, E$^2$OAL attains superior accuracy with only marginal additional cost.

\textbf{Ablation on $\hat{u}$ Estimation.} 
Figure~\ref{fig:xxyy} (Appendix) visualizes how the inferred number of unknown classes $\hat{u}$ evolves across rounds.
Even without access to class priors, our adaptive class estimation module exhibits stable convergence and produces estimates that remain within the correct order of magnitude under varying mismatch ratios.

\textbf{Ablation on Pretrained Representations.} 
Table~\ref{tab:clip} (Appendix) reports the final-round performance when the default CLIP-extracted features in Algorithm~\ref{alg:ternary_search_estimation} are replaced with representations from a pretrained MoCo~\cite{he2020momentum} backbone.
The results show that model accuracy remains largely unchanged, indicating that E$^2$OAL is robust to the choice of pretrained representation.
This consistency suggests that the effectiveness of our adaptive estimation strategy primarily arises from the proposed algorithmic design rather than dependence on a specific feature backbone.
In practice, CLIP can be substituted with any pretrained model capable of providing high-quality, task-agnostic representations.

%% file: sec/5_conclusion.tex
\section{Conclusion}
In this paper, we presented E$^2$OAL, an effective and efficient framework for open-set active learning (OSAL) that reduces wasted annotation cost while significantly improving model performance under open-set conditions.
Unlike prior approaches that depend on separate detectors or neglect the value of labeled \texttt{unknowns}, E$^2$OAL leverages them through a two-stage pipeline: (1) an adaptive class estimation and utilization module that uncovers latent structures of labeled \texttt{unknowns} in a frozen semantic space and leverages them to enhance known-class discrimination; and (2) a flexible query strategy that jointly balances informativeness and purity through Dirichlet-based calibration, logit-margin scoring, and a mild uncertainty metric tailored for open-set scenarios. Extensive experiments across multiple OSAL benchmarks demonstrate that E$^2$OAL consistently surpasses prior methods in both accuracy and query effectiveness, while requiring fewer hyperparameters and lower training overhead. Our analysis further highlights the overlooked potential of labeled \texttt{unknowns} as a valuable source of supervisory signal for open-world learning.

%% file: sec/X_suppl.tex
\clearpage
\setcounter{page}{1}
\maketitlesupplementary

\makeatletter
\renewcommand{\thefigure}  {\Alph{section}\arabic{figure}}
\renewcommand{\thetable}   {\Alph{section}\arabic{table}}
\renewcommand{\thealgorithm}{\Alph{section}\arabic{algorithm}}
\@addtoreset{figure}{section}   
\@addtoreset{table}{section}
\@addtoreset{algorithm}{section}
\makeatother

\section{Full Expression of Equation (4)}
\label{fe_eq4}

The KL divergence term $\mathcal{L}_{\mathrm{KL}}$ between the predicted Dirichlet distribution and the uniform prior can be derived as:
\begin{equation}
\begin{aligned}
&\mathcal{L}_{\mathrm{KL}} = \mathrm{KL}\big(\mathrm{Dir}(\tilde{\boldsymbol{\alpha}}) \parallel \mathrm{Dir}(\mathbf{1})\big) \\
&= \int \mathrm{Dir}(\boldsymbol{p}; \tilde{\boldsymbol{\alpha}}) \log \frac{\mathrm{Dir}(\boldsymbol{p}; \tilde{\boldsymbol{\alpha}})}{\mathrm{Dir}(\boldsymbol{p}; \mathbf{1})} d\boldsymbol{p} \\
&= \int \left(\frac{1}{\mathbb{B}\left ( \boldsymbol{\tilde{\alpha}} \right )   }{ \prod_{i=1}^{k+\hat{u}}} {p}_{i}^{\tilde{\alpha}_{i}-1}\right) \log \left(\frac{\mathbb{B}(\boldsymbol{1})}{\mathbb{B}\left(\boldsymbol{\tilde{\alpha}}\right)} \prod_{i=1}^{k+\hat{u}} {p}_{i}^{\tilde{\alpha}_{i}-1}\right) d \boldsymbol{p}\\
&= \log \frac{\mathbb{B}(\boldsymbol{1})}{\mathbb{B}\left(\boldsymbol{\tilde{\alpha}}\right)} \int\frac{1}{\mathbb{B}\left(\boldsymbol{\tilde{\alpha}}\right)} \prod_{i=1}^{k+\hat{u}} p_{i}^{\tilde{\alpha}_{ i}-1} d \boldsymbol{p} \\&+ \int\left(\log \prod_{i=1}^{k+\hat{u}} p_{i}^{\tilde{\alpha}_{i}-1}\right)\left(\frac{1}{\mathbb{B}\left(\tilde{\alpha}\right)} \prod_{i=1}^{k+\hat{u}} p_{i}^{\tilde{\alpha}_{i}-1}\right) d \boldsymbol{p} \\
&= \log \frac{\mathbb{B}(\boldsymbol{1})}{\mathbb{B}\left(\boldsymbol{\tilde{\alpha}}\right)}+\mathbb{E}_{\boldsymbol{p} \sim \mathrm{Dir}(\boldsymbol{\tilde{\alpha}})} \left[\log \textstyle \prod_{i=1}^{k+\hat{u}} p_{i}^{\tilde{\alpha}_{i}-1}\right]\\
&= \log \frac{\mathbb{B}(\boldsymbol{1})}{\mathbb{B}\left(\boldsymbol{\tilde{\alpha}}\right)}+\sum_{j=1}^C\left(\tilde{\alpha}_{ j}-1\right) \mathbb{E}_{\boldsymbol{p}_j \sim {\mathcal{B} }\left(\tilde{\alpha}_{j},  {\sum_{i \ne j}} \tilde{\alpha} _{i}\right)} \left[\log p_{j} \right] \\
& = \log \left[\frac{\Gamma\left(\sum_{i=1}^{k+\hat{u}} \tilde{\alpha}_{i}\right)}{\Gamma({k+\hat{u}}) \prod_{i=1}^{k+\hat{u}} \Gamma\left(\tilde{\alpha}_{ i}\right)}\right]\\&+\sum_{j=1}^{k+\hat{u}}\left(\tilde{\alpha}_{ j}-1\right)\left[\psi\left(\tilde{\alpha}_{j}\right)-\psi\left(\sum_{i=1}^{k+\hat{u}} \tilde{\alpha}_{i}\right)\right],
\end{aligned}
\end{equation}
where $\mathbb{B}(\cdot)$ denotes the multivariate Beta function,  
$\mathcal{B}(\cdot, \cdot)$ is the standard Beta function,  
$\Gamma(\cdot)$ represents the Gamma function, and $\psi(\cdot)$ denotes the digamma function. The explicit definitions of $\mathbb{B}(\cdot)$ and $\mathcal{B}(\cdot,\cdot)$ are given by: 
\begin{equation}
\mathbb{B}(\tilde{\boldsymbol{\alpha}}) = \frac{ \prod_{i=1}^{k+\hat{u}} \Gamma(\tilde{\alpha}_i)}{\Gamma\left( \sum_{i=1}^{k+\hat{u}} \tilde{\alpha}_i \right)},
\end{equation}
and
\begin{equation}
{\mathcal{B} }\left(\tilde{\alpha}_{j},  {\sum_{i \ne j}} \tilde{\alpha} _{i}\right)=\frac{\Gamma\left (\tilde{\alpha}_{j}  \right ) \Gamma\left ({\sum_{i \ne j}} \tilde{\alpha} _{i} \right )}{\Gamma\left (  \tilde{\alpha}_{j}+  {\sum_{i \ne j}} \tilde{\alpha} _{i}\right ) } .
\end{equation}

\section{The Pseudocode of E²OAL}
We outline the full workflow of E$^2$OAL in Algorithm~\ref{alg:active_learning_query}, which covers each stage of the active learning process.

\begin{algorithm*}
\caption{The E$^2$OAL algorithm}
\label{alg:active_learning_query}
\textbf{Input:} Labeled pool $\mathcal{D}_{L} = \mathcal{D}_{L}^{kno}\cup \mathcal{D}_{L}^{unk}$, unlabeled pool $\mathcal{D}_{U}$, known class count $k$, upper limit $\hat{u}_{max}$, target classifier $f_{\theta}$, query budget $|\mathcal{B}|$, and target query precision $p^*$.\\
\textbf{Process: (The $t$-th active learning round)}

\begin{algorithmic}[1]
    \IF{$t = 1$}
    \STATE \textit{\# Model training}
    \STATE Train classifier $f_\theta$ using $\mathcal{L}_{\mathrm{CE}}$ for primary head ($k$-way) and $\mathcal{L}_{\mathrm{EDL}}$ for auxiliary head ($k$-way)

    \STATE \textit{\# Purity-based candidate selection}
    \STATE Obtain auxiliary head's outputs for all $x \in \mathcal{D}_L \cup \mathcal{D}_U$
    \FOR{each $x \in \mathcal{D}_L \cup \mathcal{D}_U$}
      \STATE Compute purity margin: $S_{\text{purity}}(x) = \max_{c \in \mathcal{C}_k} o_c$
    \ENDFOR
    \ELSE
    \STATE \textit{\# Unknown class estimation}
    \STATE Estimate $\hat{u}$ based on labeled pool $\mathcal{D}_L$, known class count $k$, and upper bound $\hat{u}_{\max}$ using Algorithm~\ref{alg:ternary_search_estimation}

    \STATE \textit{\# Model training}
    \STATE Train classifier $f_\theta$ using $\mathcal{L}_{\mathrm{CE}}$ for primary head ($k$-way) and $\mathcal{L}_{\mathrm{EDL}}$ for auxiliary head ($(k+\hat{u})$-way)

    \STATE \textit{\# Purity-based candidate selection}
    \STATE Obtain auxiliary head's outputs for all $x \in \mathcal{D}_L \cup \mathcal{D}_U$.
    \FOR{each $x \in \mathcal{D}_L \cup \mathcal{D}_U$}
      \STATE Compute purity margin: $S_{\text{purity}}(x) = \max_{c \in \mathcal{C}_k} o_c - \max_{c \in \mathcal{C}_{\hat{u}}} o_c$
    \ENDFOR
    \ENDIF
    \STATE Fit a 3-component Gaussian Mixture Model (GMM)~\cite{permuter2006study} on logit margins $\{S_{\text{purity}}(x)\}$ to model different purity regimes: high (known), low (unknown), and intermediate (ambiguous)
    \STATE For each $x \in \mathcal{D}_U$, compute its likelihood under the high-purity component
    \STATE Sort $\mathcal{D}_U$ in descending order of the computed likelihoods
    \STATE Initialize the candidate pool $\mathcal{C}_{pool}$ with the top $|\mathcal{B}|$ samples from the sorted $\mathcal{D}_U$

    \STATE \textit{\# Precision-based candidate refinement}
    \STATE Compute the calibrated target query precision $\hat{p}^*_{t} = 
    \begin{cases}
    \max\big(\min(\hat{p}^*_{t-1} + (p^* - \bar{p}_{t-1}^*), 0), 1\big) & \text{if } t > 1 \\
    p^* & \text{if } t = 1
    \end{cases}$
    \WHILE{the mean likelihood of the lowest $|\mathcal{B}|$ samples in $\mathcal{C}_{\text{pool}}$ is greater than $\hat{p}^*_t$}
        \STATE Add the next highest-likelihood sample from $\mathcal{D}_U$ into $\mathcal{C}_{\text{pool}}$
    \ENDWHILE

    \STATE \textit{\# Information-based final query selection}
    \FOR{each $x \in \mathcal{C}_{\text{pool}}$}
      \STATE Obtain predicted probability vector $\mathbf{p}$ of $x$ from the primary head
    \STATE Let $\mathbf{u}$ be a uniform distribution over all classes
    \STATE Let $\mathbf{p}^{\mathrm{max}}$ be a one-hot vector with 1 at $\arg\max(\mathbf{p})$
    \STATE Compute information score: $S_{\text{info}}(x) = \mathrm{JS}\big(\mathbf{p} \parallel \mathbf{u}\big) \cdot \mathrm{JS}\big(\mathbf{p} \parallel \mathbf{p}^{\mathrm{max}}\big)$
    \ENDFOR

    \STATE Select top $|\mathcal{B}|$ samples from $\mathcal{C}_{pool}$ with the highest $S_{\text{info}}(x)$ scores as the query set $\mathcal{B}$
    \STATE Compute observed query precision: $\bar{p}^*_t = \frac{|\mathcal{B}^{kno}|}{|\mathcal{B}|}$ 
    \STATE Update data pools: $\mathcal{D}_{U} \leftarrow \mathcal{D}_{U} \setminus \mathcal{B}$, $\mathcal{D}_{L} \leftarrow \mathcal{D}_{L} \cup \mathcal{B}$
    \STATE \textbf{return}~ $\mathcal{D}_{L}$, $\mathcal{D}_{U}$, $\bar{p}^*_t$, and $f_{\theta}$ for the next round
    
\end{algorithmic}
\end{algorithm*}

\section{Additional Results for Figure 1}

\begin{figure*}[!ht]
  \centering
  \begin{subfigure}[ht]{0.495\linewidth}
    \centering
    \includegraphics[width=\linewidth]{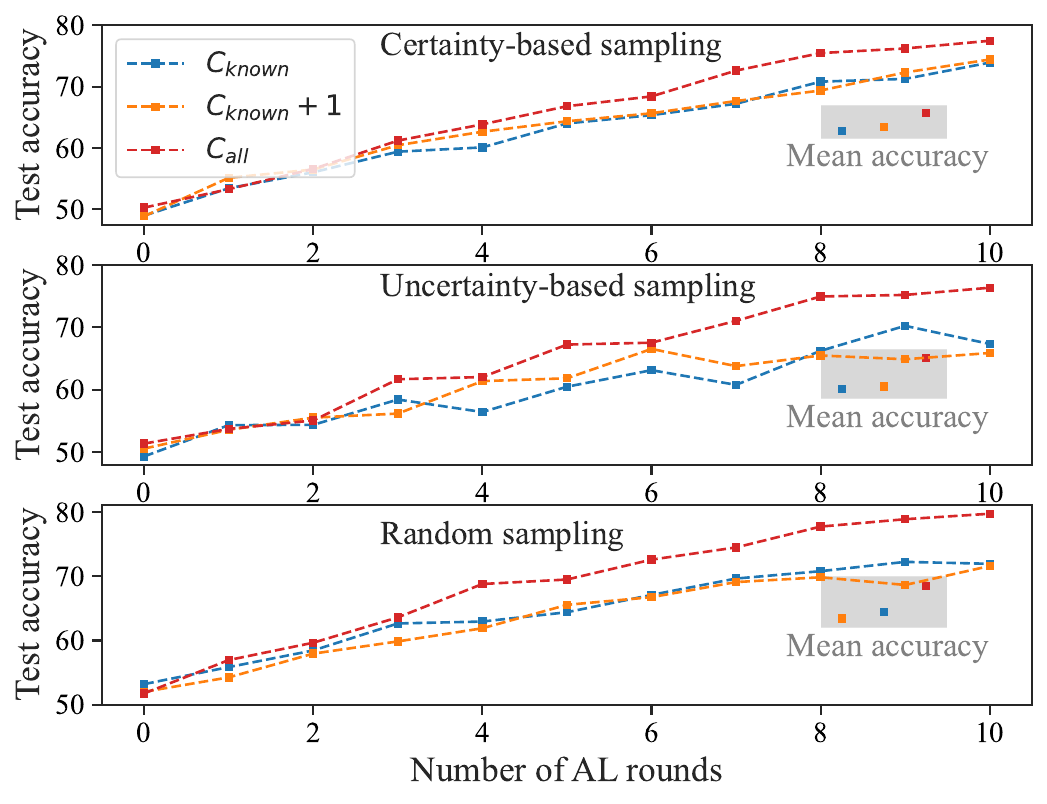}
    \caption{ResNet-18, 20 known / 80 unknown classes}
  \end{subfigure}
  \begin{subfigure}[ht]{0.495\linewidth}
    \centering
    \includegraphics[width=\linewidth]{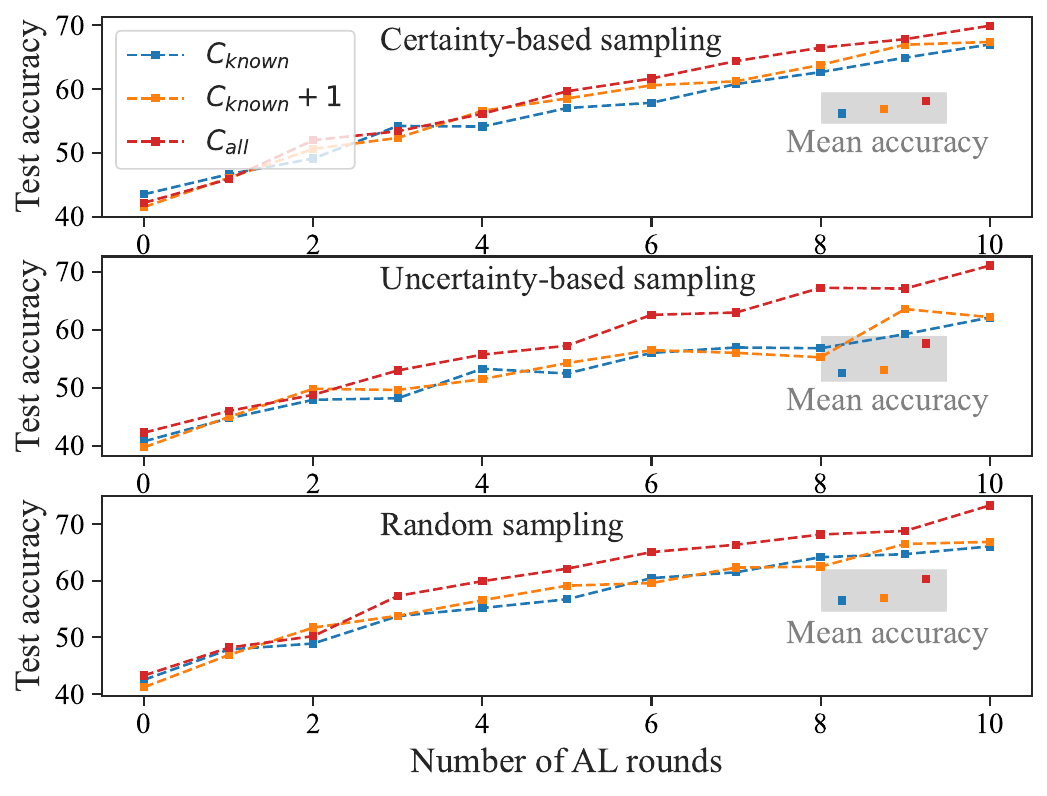}
    \caption{ResNet-18, 30 known / 70 unknown classes}
  \end{subfigure}
  \vskip\baselineskip
  \begin{subfigure}[ht]{0.495\linewidth}
    \centering
    \includegraphics[width=\linewidth]{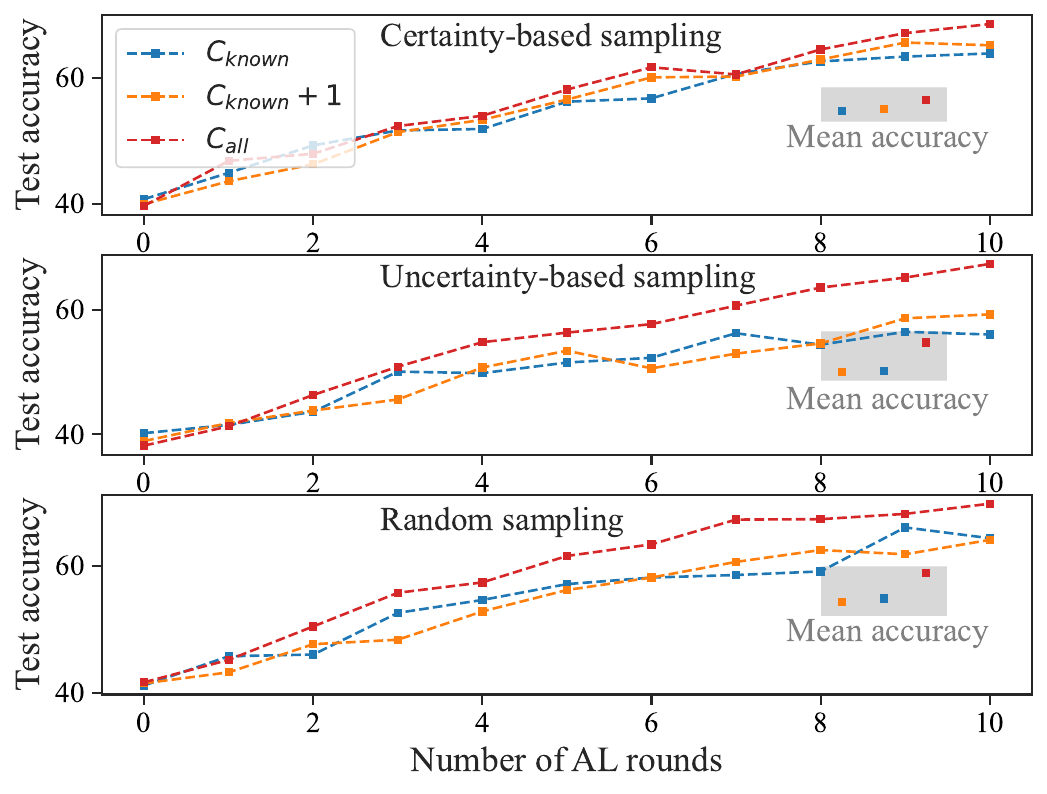}
    \caption{ResNet-34, 30 known / 70 unknown classes}
  \end{subfigure}
  \begin{subfigure}[ht]{0.495\linewidth}
    \centering
    \includegraphics[width=\linewidth]{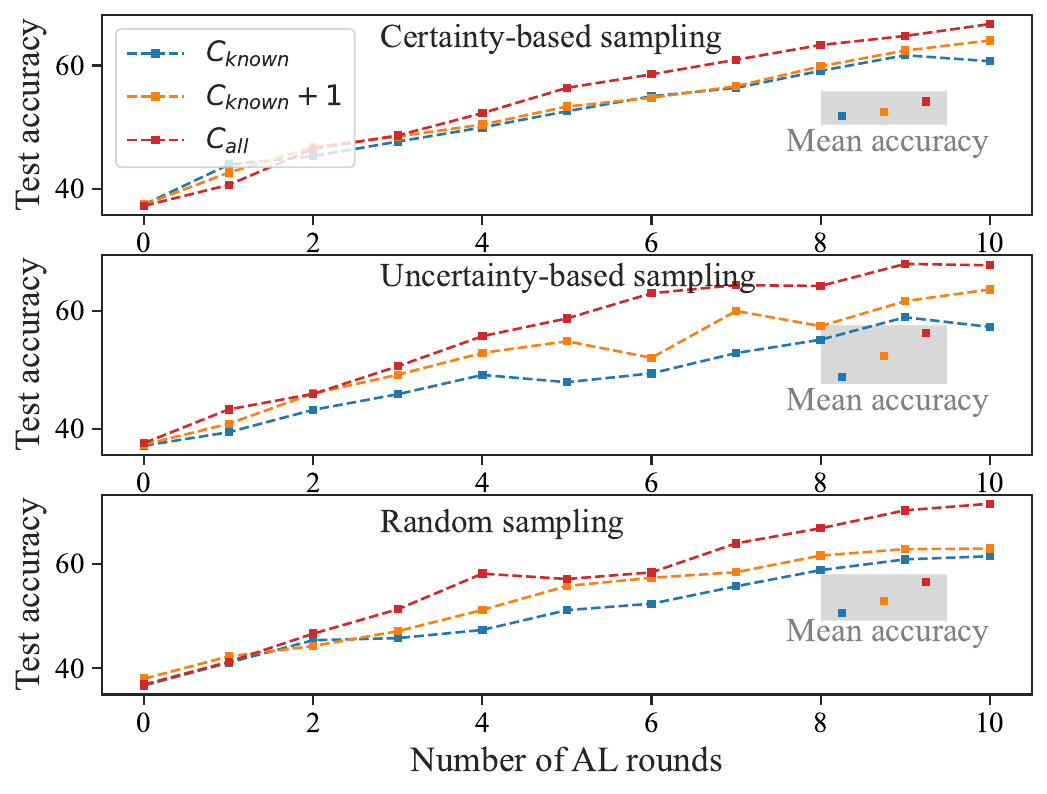}
    \caption{ResNet-50, 30 known / 70 unknown classes}
  \end{subfigure}
  \caption{Per-round and mean test accuracy on CIFAR-100 under varying class splits and network architectures. $C_{known}$ excludes labeled \texttt{unknowns}, $C_{known{+}1}$ collapses them into a single class, and $C_{all}$ leverages their ground-truth labels.}
  \label{fig:motivation_combined}
\end{figure*}

We present additional results over a wider range of mismatch ratios and network architectures beyond those in Figure~\ref{fig:motivation}. As illustrated in Figure~\ref{fig:motivation_combined}, we evaluate three sampling strategies to assess whether labeled \texttt{unknowns} can enhance known-class learning: (1) random sampling; (2) certainty-based sampling via maximum softmax probability (MSP~\cite{hendrycks2016baseline}); and (3) uncertainty-based sampling using the least-confidence criterion~\cite{li2006confidence}. Our observations indicate that, regardless of sampling strategy, mismatch ratio, or network architecture, leveraging fine-grained labels of labeled \texttt{unknowns} within the auxiliary classifier consistently yields the best performance—especially as network capacity increases. Treating all labeled \texttt{unknowns} as a single class typically improves over ignoring them but results in less stable and less substantial gains, occasionally even degrading performance. These findings highlight the significant benefits of effectively exploiting labeled \texttt{unknowns} during training, motivating the in-depth study and method proposed in this work.

\section{Additional Results for Figure 3}

\begin{figure*}[!ht]
\centering
   \includegraphics[width=1.\linewidth]{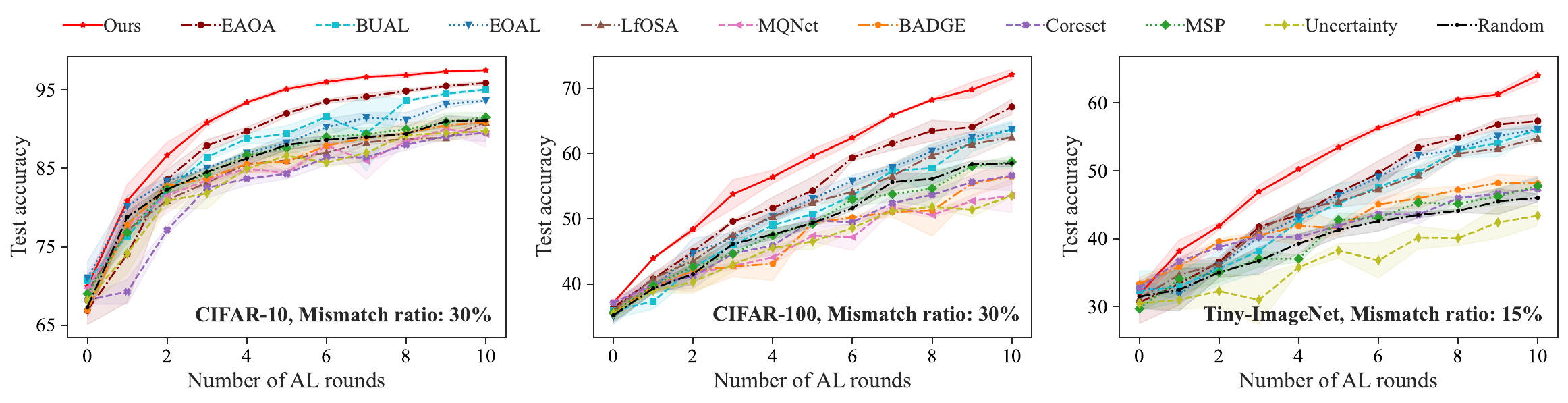}
\caption{Test accuracy across rounds under mismatch ratio 30\% on CIFAR-10/100 and 15\% on Tiny-ImageNet.}
	\label{fig:acc_curve_appendix}
\end{figure*}

Figure~\ref{fig:acc_curve_appendix} illustrates the evolution of test accuracy across rounds under intermediate mismatch ratios on three benchmark datasets.
E$^2$OAL consistently surpasses all baselines, exhibiting a clearly superior accuracy trajectory throughout the process.
Notably, the performance gap widens as the dataset complexity increases, highlighting the robustness and scalability of the proposed framework.

\section{Additional Results for Figure 4}

\begin{figure*}[!ht]
\centering
   \includegraphics[width=0.95\linewidth]{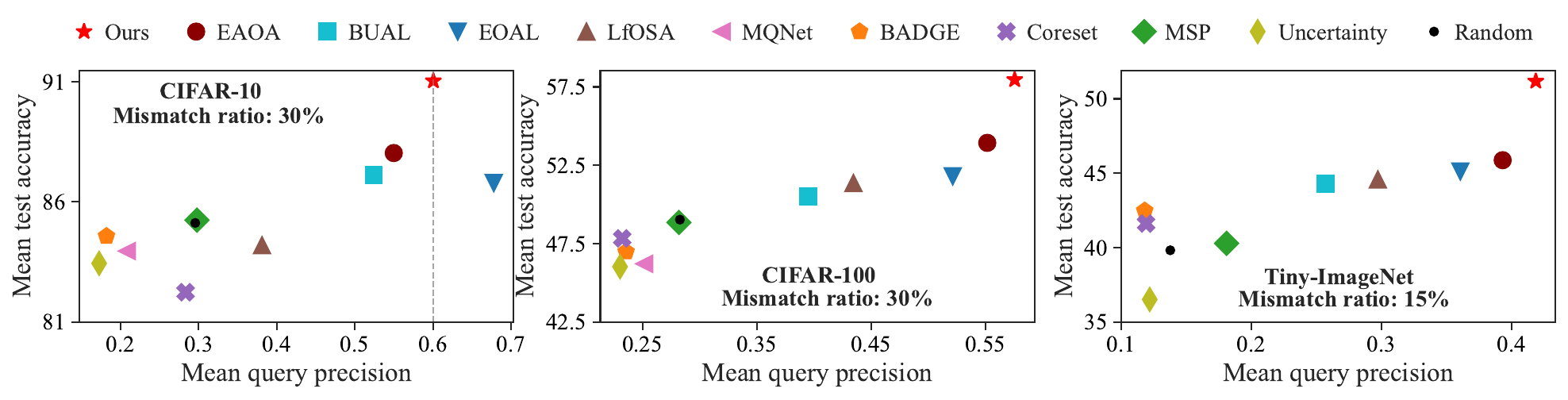}
\caption{Mean query precision and test accuracy under mismatch ratio 30\% on CIFAR-10/100 and 15\% on Tiny-ImageNet.}
	\label{fig:query_appendix}
\end{figure*}

Figure~\ref{fig:query_appendix} reports the mean test accuracy and average query precision across rounds under moderate mismatch ratios on three benchmark datasets.
E$^2$OAL achieves the highest query precision on CIFAR-100 and Tiny-ImageNet, while consistently outperforming all competing methods in accuracy.
On CIFAR-10, the query precision is slightly lower than EOAL~\cite{safaei2024entropic}, which is expected since our purity control is explicitly regulated by a target precision of 0.6.
This observation underscores two key insights:
(1) Simply maximizing query purity is not optimal for open-set active learning—although EOAL attains high precision, its queried samples, albeit from known classes, tend to be less informative, leading to suboptimal accuracy;
(2) Although E$^2$OAL does not achieve the absolute highest query precision, it adheres more closely to the target value than EAOA~\cite{zong2025rethinking}, which is also guided by the target precision, demonstrating superior control and stability in purity regulation.

\begin{figure*}[!ht]
\centering
   \includegraphics[width=0.96\linewidth]{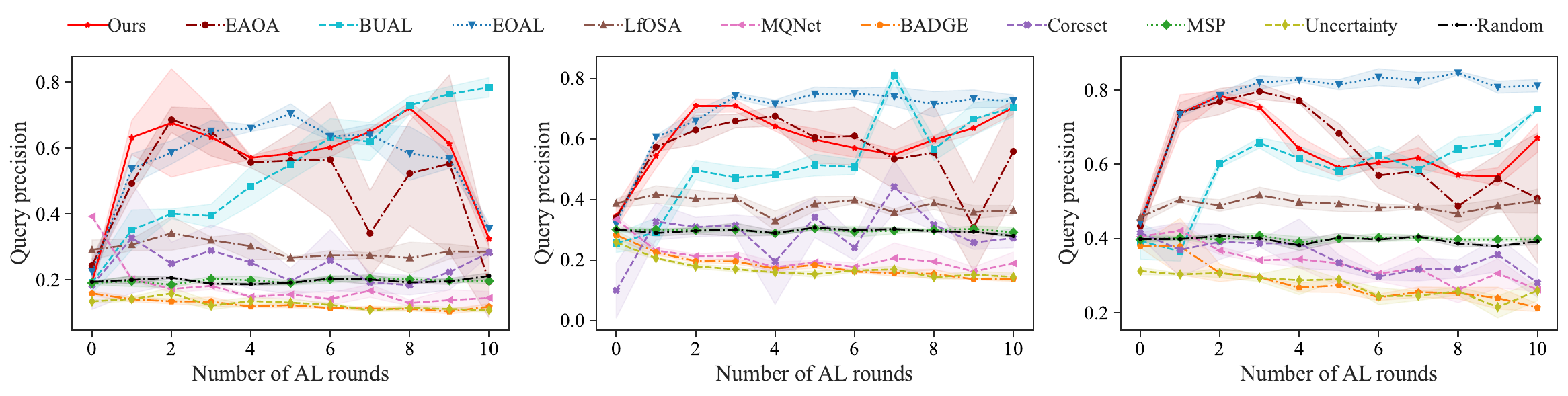}
   \includegraphics[width=0.96\linewidth]{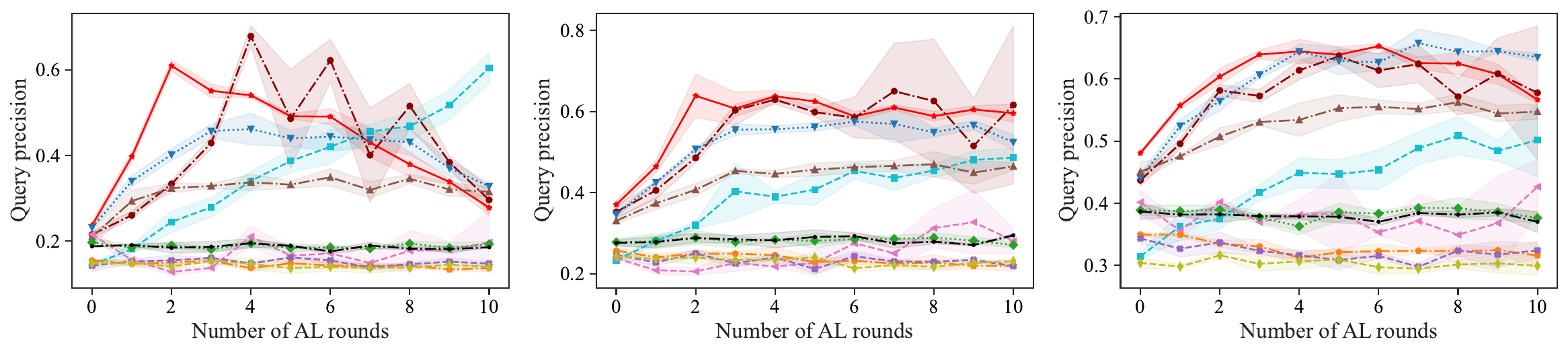}
   \includegraphics[width=0.96\linewidth]{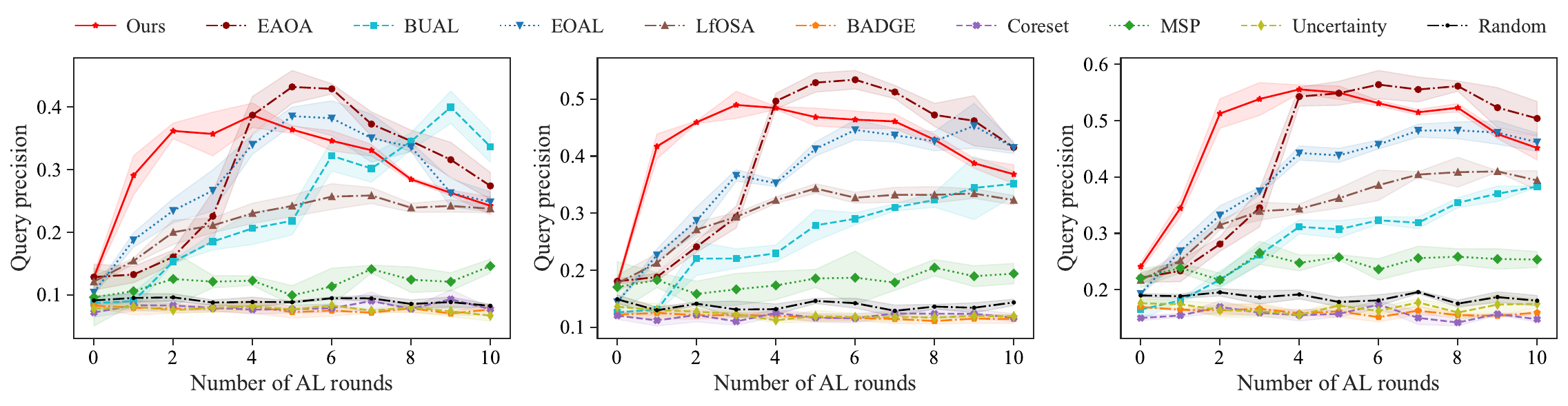}
\caption{Cross-round query precision under varying mismatch ratios on three benchmarks. From top to bottom: results on CIFAR-10, CIFAR-100, and Tiny-ImageNet; from left to right: mismatch ratios of 20\%, 30\%, and 40\%, respectively.}
	\label{fig:query_curve_appendix}
\end{figure*}

Figure~\ref{fig:query_curve_appendix} further illustrates how query precision evolves over rounds. It can be observed that our method maintains a high query precision even in the early training stages—a desirable property, as highlighted in \cite{park2022meta}, where high-purity samples tend to provide greater utility when the model is still undertrained. This advantage stems from our proposed purity metric and flexible sampling strategy, which effectively constructs high-purity candidate sets. In contrast, the previous state-of-the-art method, EAOA, shows suboptimal query precision in the early rounds and exhibits significant fluctuations, especially on CIFAR-100 with a 20\% mismatch ratio. This instability arises from its inherent limitation in the adaptive sampling strategy, where the adjustment of the parameter $k$ is limited to a fixed step size per round, hindering rapid convergence to an optimal value. Our method, by comparison, introduces no additional hyperparameters and achieves stable, high-precision queries from the outset, demonstrating clear superiority.

\section{Additional Results for Table 1}

\begin{table*}[!ht]
\centering
\begin{tabular}{lccccccccc}
 \toprule
\textbf{Dataset} & \multicolumn{3}{c}{\textbf{CIFAR-10}} & \multicolumn{3}{c}{\textbf{CIFAR-100}} & \multicolumn{3}{c}{\textbf{Tiny-ImageNet}} \\
 \midrule
\textbf{Mismatch ratio} & 20\% & 30\% & 40\% & 20\% & 30\% & 40\% & 10\% & 15\% & 20\% \\
 \midrule
Random & 94.48 & 91.11 & 87.18 & 64.45 & 58.48 & 55.48 & 47.93 & 46.00 & 43.32  \\
 \midrule
Uncertainty~\cite{li2006confidence}    & 95.70 & 89.77 & 83.61 & 62.25 & 53.52 & 50.83 & 45.83 & 43.40 & 35.43  \\
Coreset~\cite{sener2017active} & 94.20 & 89.56 & 86.38 & 63.53 & 56.62 & 55.00 & 50.60 & 47.33 & 45.35  \\
BADGE~\cite{ash2019deep}      & 94.95 & 90.91 & 87.12 & 64.00 & 56.49 & 50.20 & 49.70  & 48.16 & 46.23 \\
 \midrule
MSP~\cite{hendrycks2016baseline}   & 94.15 & 91.51 & 87.21 & 65.33 & 58.69 & 56.68 & 51.43 & 47.78 & 46.57 \\
LfOSA~\cite{ning2022active}    & 94.15 & 90.91 & 87.43 & 70.32 & 62.49 & 58.49 & 58.37 & 54.78 & 51.33 \\
 \midrule
 MQNet~\cite{park2022meta}   & 95.12 & 89.39 & 87.42 & 63.70 & 53.52 & 55.44 & - & - & - \\
EOAL~\cite{safaei2024entropic}    & 96.23 & 93.64 & 91.63 & 73.73 & 63.69 & 59.55 & 61.40 & 56.13 & 52.65 \\
 BUAL~\cite{zong2024bidirectional}   & 96.48 & 95.04 & 92.52 & 73.43 & 63.73 & 59.89 & 63.80 & 56.09 & 50.52 \\
EAOA~\cite{zong2025rethinking}    & 97.23 & 95.88 & 93.09 & 74.60 & 67.14 & 63.49 & 62.33 & 57.31 & 53.33 \\
 \midrule
\textbf{Ours*} & \underline{97.33} & \underline{95.94} & \underline{93.13} & \underline{75.90} & \underline{67.54} & \underline{63.85} & \underline{64.23} & \underline{60.44} & \underline{54.73}  \\
$\uparrow$ over best baseline (\%) & 0.10 & 0.06 & 0.04 & 1.30 & 0.40 & 0.36 & 1.15 & 3.13 & 1.40  \\
\midrule
\textbf{Ours} & \textbf{98.77} & \textbf{97.52} & \textbf{95.69} & \textbf{82.20} & \textbf{72.10} & \textbf{67.98} & \textbf{68.53} & \textbf{64.02} & \textbf{57.10}  \\
$\uparrow$ over best baseline (\%) & 1.44 & 1.64 & 2.60 & 7.60 & 4.96 & 4.49 & 4.73 & 6.71 & 3.77  \\
\bottomrule
\end{tabular}
\caption{Final-round test accuracy (\%) of all methods under varying mismatch ratios on CIFAR-10, CIFAR-100, and Tiny-ImageNet. ``Ours*'' denotes a variant of our method where the target classifier is trained independently without leveraging labeled \texttt{unknowns}. The best result in each setting is highlighted in bold, while the second best is underlined. Due to the poor performance and high training cost of MQNet, we do not include it on Tiny-ImageNet, and thus mark it with ``–".}
\label{tab:final-acc}
\end{table*}

Table~\ref{tab:final-acc} summarizes the final-round test accuracy of all methods under different mismatch ratios across three benchmark datasets.
We report results for both ``Ours" (the proposed E$^2$OAL) and ``Ours*" (a variant where the target classifier is trained independently without utilizing labeled \texttt{unknowns}, similar to prior baselines).
Both E$^2$OAL and its variant consistently outperform existing methods by a notable margin.
The strong performance of ``Ours*'' demonstrates the effectiveness of our adaptive sampling strategy in selecting more informative known-class samples, while the additional improvement achieved by E$^2$OAL further highlights the value of leveraging labeled \texttt{unknowns} to enhance known-class learning.

\section{Ablation on CLIP Representations}
\label{CLIP}

\begin{table}[!ht]
\centering
\small
\begin{tabular}{lccc}
\toprule
 Feature source & CIFAR-10 & CIFAR-100 & Tiny-ImageNet \\
\midrule
CLIP      & \textbf{97.52} & 72.10 & \textbf{64.02} \\
MoCo     & 97.44 & \textbf{72.31} & 63.87 \\
\bottomrule
\end{tabular}
\caption{Final-round test accuracy (\%) under fixed mismatch ratios (30\% for CIFAR-10/100 and 15\% for Tiny-ImageNet).
``CLIP" and ``MoCo" respectively refer to adaptive class estimation performed using features extracted from a CLIP model and from a self-supervised MoCo pretrained model.
}
\label{tab:clip}
\end{table}

Table~\ref{tab:clip} presents the final-round results when the default CLIP features are replaced with self-supervised MoCo pretrained representations for adaptive class estimation.
E$^2$OAL exhibits stable performance across all datasets, with only marginal differences between the two feature sources.
These results demonstrate that E$^2$OAL is robust to the choice of pretrained representation, and its effectiveness does not depend on a specific feature extractor.
In practice, CLIP can be substituted with any pretrained model capable of providing high-quality, task-agnostic features.

\section{Ablation on û Estimation}
\label{u-hat}

\begin{figure*}[!ht]
\centering
   \includegraphics[width=1.\linewidth]{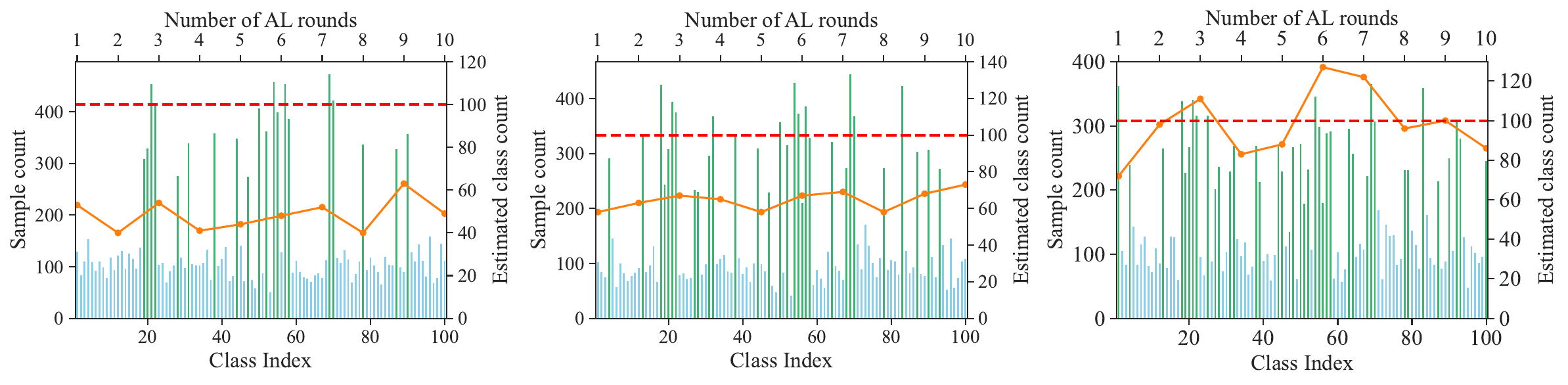}
\caption{Ablation results for unknown class estimation on CIFAR-100 under mismatch ratios of 20\%, 30\%, and 40\% (from left to right). Bar charts (bottom x-axis and left y-axis) show the total number of samples labeled per class in the final round. Green bars represent known classes, and blue bars represent unknown classes. Line plots (top x-axis and right y-axis) illustrate the evolution of the estimated total number of classes ($k + \hat{u}$), where the ground-truth total is 100.}
	\label{fig:xxyy}
\end{figure*}

Figure~\ref{fig:xxyy} illustrates the estimated total number of classes, $k + \hat{u}$, across rounds on CIFAR-100 under mismatch ratios of 20\%, 30\%, and 40\%, where $k$ denotes the number of known classes (20, 30, and 40, respectively).
For reference, the figure also includes the final number of queried samples per class.
Our adaptive class estimation module consistently yields stable and reliable estimates across rounds and mismatch settings.
Since the true class prior of unknown samples is inaccessible, the estimated total number of classes may not exactly match the ground truth.
This deviation partly arises from the inherent ambiguity of class granularity—for instance, CIFAR-100 can be organized into either 20 coarse-grained or 100 fine-grained categories.
Nevertheless, our method consistently provides estimates within the correct order of magnitude, with accuracy improving as the mismatch ratio increases (i.e., as the open-set problem becomes less challenging).
In particular, under the 40\% setting, the estimated class count closely fluctuates around the true value, highlighting the effectiveness and robustness of the proposed class estimation strategy.